\documentclass{article}

\usepackage{microtype}
\usepackage{graphicx}
\usepackage{booktabs} 

\usepackage{amsmath}
\usepackage{amssymb}
\usepackage{pifont}
\usepackage[normalem]{ulem}
\usepackage{multirow}

\usepackage{times}
\usepackage{subcaption} 

\usepackage{natbib}

\usepackage{algorithm}
\usepackage{algorithmic}

\usepackage{hyperref}

\usepackage[accepted]{icml2018}

\icmltitlerunning{Detecting Anomalous Faces with `No Peeking' Autoencoders}

\begin{document} 
\twocolumn[
\icmltitle{Detecting Anomalous Faces with `No Peeking' Autoencoders}




\begin{icmlauthorlist}
\icmlauthor{Anand Bhattad}{UIUC}
\icmlauthor{Jason Rock}{UIUC}
\icmlauthor{David Forsyth}{UIUC}
\end{icmlauthorlist}

\icmlaffiliation{UIUC}{University of Illinois, Urbana-Champaign}

\icmlcorrespondingauthor{Anand Bhattad}{bhattad2@illinois.edu}
\icmlkeywords{Auto Encoders, Anomaly Detection, Inpainting}


\vskip 0.3in
]



\printAffiliationsAndNotice{}  

\begin{abstract} 
Detecting anomalous faces has important applications.  For example, a system
might tell when a train driver is incapacitated by a medical event, and assist in adopting
a safe recovery strategy.  These applications are demanding, because they require
accurate detection of rare anomalies that may be seen only at runtime. 
Such a setting causes supervised methods to perform poorly. We describe a method for detecting an 
anomalous face image that meets these requirements.   
We construct a feature vector that reliably has large entries for anomalous
images, then use various simple unsupervised methods to score the image
based on the feature. Obvious
constructions (autoencoder codes; autoencoder residuals) are defeated by a
`peeking' behavior in autoencoders.    Our feature construction
removes rectangular patches from the image, predicts the likely content of the patch conditioned
on the rest of the image using a specially trained autoencoder, then compares
the result to the image.    High scores suggest that the patch was difficult for an autoencoder
to predict, and so is likely anomalous.
We demonstrate that our method can identify real anomalous face images in
pools of typical images, taken from celeb-A,  that is much larger than
usual in state-of-the-art experiments. A control experiment based on our method with 
another set of normal celebrity images - a `typical set', but non-celeb-A are not 
identified as anomalous; confirms this is not due to special properties of celeb-A.
\end{abstract} 

\section{Introduction}
\label{introduction}
  
We describe a method for detecting anomalous faces in images.  Our method uses a novel representation of appearance
(auto-encoder residuals), and does not require any example anomaly in training.  We demonstrate that our method
significantly improves over a number of natural baselines.

Detecting anomalous faces has important applications.  For instance, a machine operator might fall asleep or have a heart
attack.  Ideally, a monitoring system would identify this kind of problem by watching the operator's face and trigger
some form of intervention. The crucial difficulty in building such a system is that there aren't datasets showing
(say) people having heart attacks. Moreover, a reliable anomaly detection system must be built without seeing actual
anomalies to generalize well.

This example presents serious difficulties for current methods for anomaly detection (briefly reviewed below), because
previous anomaly detection systems tend to be evaluated on datasets where anomalies are very different from typical
examples. But anomalous faces look quite similar to typical faces. Our method requires a representation of 
face appearance which exaggerates the relatively small changes that make a face image anomalous, without actually being
shown.  Worse, because face images are relatively high dimensional, there is no practical prospect of
simply applying a density estimator to the example images.  
Our strategy is to learn a compression procedure that reconstructs faces well, but not other similar unseen images, 
and then look at the residuals. This is not a routine application because one must be sure that (a) training images 
reconstruct well (routine) but (b) other similar images do not (tricky, and unusual). We show that a carefully designed 
residual of a specially trained autoencoder has these two properties and therefore provides a strong feature for identifying facial anomalies.
  
\noindent\textbf{Contributions:} We augment the Celeb-A dataset \cite{liu2015faceattributes} for evaluating image anomaly detection.
We present a novel feature learning approach for anomaly detection using inpainting auto-encoders.
We build a dataset of real anomalous faces and real typical faces to evaluate the proposed framework.
We demonstrate that our feature works well in both supervised and unsupervised applications.

\section{Background}
\noindent\textbf{Anomaly detection} has widespread applications, including: image matting \cite{Hasler:2003dz}; identifying
cancerous tissue \cite{Alpert:2014bz};  finding problems in textiles \cite{Serdaroglu:2006he,Mak:2005gy}; 
and preventing face spoofing \cite{Arashloo:2017ge}.  There is a recent survey 
in~\cite{Chandola:2009fo}.    There are two distinct types of approach in the literature.  In one approach, examples of both
inliers and outliers are available, and discriminative procedures can be used to build representations and
identify and select features.  In the other, one can model only inliers, and anomalies are available only at test time.

\textbf{Face anomaly detection}
is a good example problem because (a) data resources of typical faces are abundant and (b) anomalous faces look a lot like typical
faces; trivial methods perform poorly.  We do not assume that anomalous faces are available at training time, because doing so creates two problems.
First, anomalous face images are rare (which is why they're anomalous) and highly variable in appearance, so a dataset
of reasonable size is difficult to build.  Second, the estimate of the decision boundary produced by any particular set
of anomalous face images is  likely to be inaccurate. The location of the decision boundary is determined by both the
anomalies and the typical images; but the anomalies must be severely undersampled, and so contribute significant
variance to the estimate of the decision boundary.   

Instead, we assume that only typical faces are available at training time.  We must now build some form
of distribution model for true faces and exploit it to tell how uncommon the current image is.   We focus on building a feature
construction that allows simple mechanisms to compute an anomaly score.   An alternative is to use a
kernel method to build a distribution model (the \textbf{one-class SVM} of \cite{scholkopf2000support}).  We use this
method as a baseline.

Our feature construction uses an \textbf{autoencoder} \cite{hinton2006reducing}.
Auto encoders use an encoder to compress a signal to a code, which can then be decompressed.  The code is a low dimensional
representation of content which has been shown to be useul for tasks such as: appearance editing~\cite{AttributeImageCon:Cqjisrcl,Lample:2017tw}; inpainting 
\cite{Context:CktiXfRg}; and colorization \cite{Deshpande:2017ik}.  Generative adversarial networks (GAN)~\cite{Goodfellow:2014wp}
have been used for anomaly detection in~\cite{schlegl:2017au}, but one must build a distribution for the code.  
\cite{schlegl:2017au} do explore using residuals in combination with the code likelihood.  However, because their model is built on
a GAN, their inference procedure is quite expensive, requiring many backprop and gradient steps, while our method is
simply a forward run through an autoencoder.  Our model also introduces a novel inpainting conditioning strategy for feature
construction.

Evaluating anomaly detectors is tricky, because anomalies are rare.  One strategy is to regard one class of image
as typical, and another  as anomalous.  This strategy is popular
\cite{zhai2016deep, kliger2018novelty,deecke2018anomaly,gu2018semisupervised} but may mislead.  The danger is that one may
unknowingly work with two very different classes, meaning that the quality of the distribution model for the typical
class is not  tested.    In contrast, face anomaly detection has the advantage of being (a) intrinsically useful and (b)
clearly difficult. 

The set building method of \cite{Zaheer:2017wn} could be applied to face anomaly detection.  This approach
has been shown to be accurate at identifying the one special face in a set of 16.  A direct comparison is not possible,
because their method relies on identifying the one different face in a set (i.e. given 15 smiling faces and one
frowning face, it should mark the frowning face).  However, we adopt their evaluation methodology and use analogous
scoring methods.

\section{Anomaly Features}

\begin{figure*}[t]
\centering
\includegraphics[width=\linewidth]{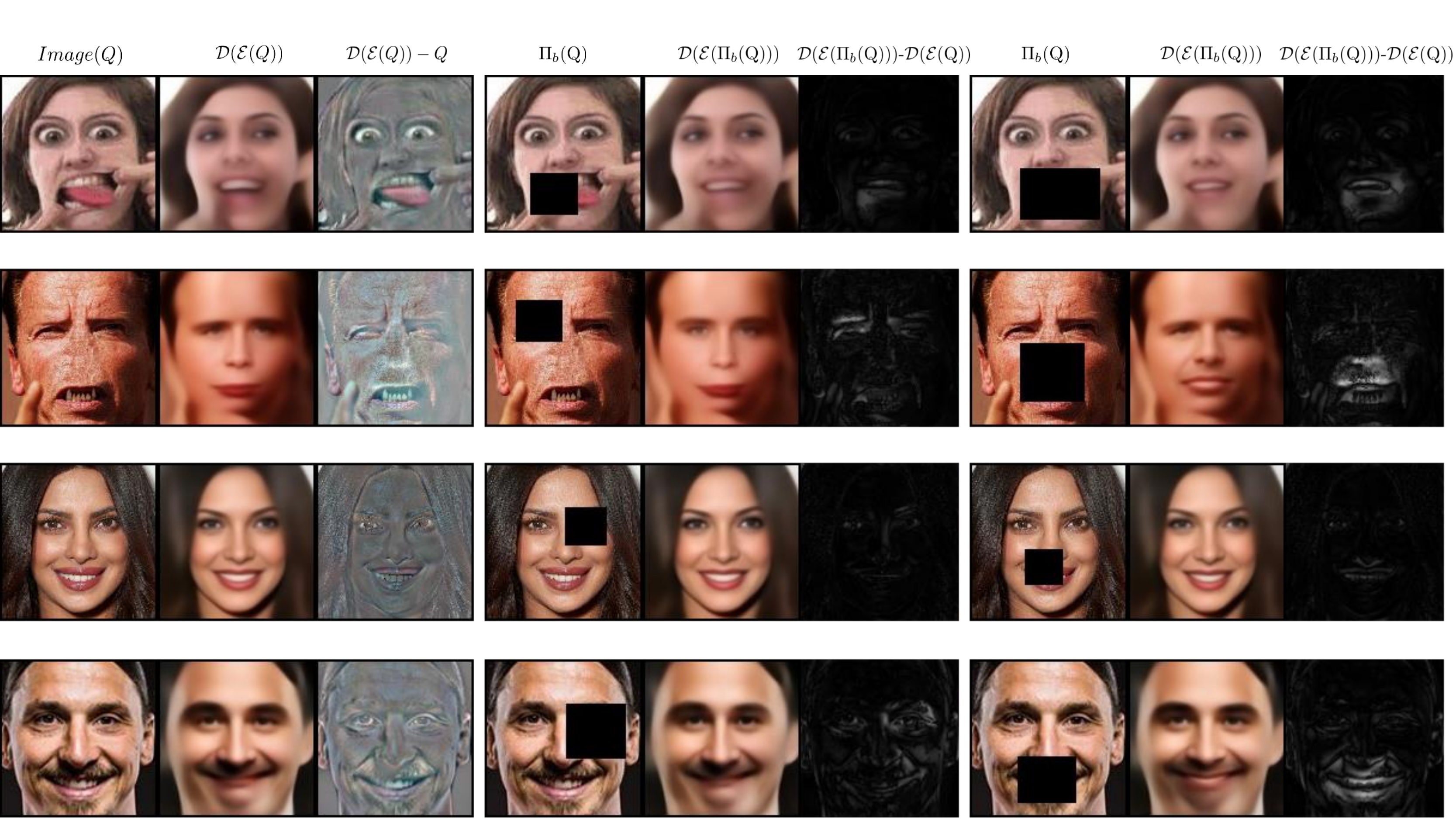}
\caption{
Forcing an autoencoder to inpaint at test time has important effects on the reconstruction.  {\bf Top two rows:} anomalous face images; {\bf bottom two rows} typical face images. 
In the {\bf first column}, the three images are actual input images, autoencoder's reconstructed images and the autoencoder residuals respectively. In the {\bf second} and {\bf third column}, 
the first image is the masked input, second is the autoencoder's reconstructed images  and third is the residual difference between inpainted reconstruction residual from the residuals in the the first column.
Notice how, for anomalous faces, not showing the autoencoder the content of the box affects the reconstruction.  In the top row, attend to the dark bar at the left side of the model's
mouth, significantly reduced when the autoencoder reconstructs without seeing Q (i.e. no peeking).  Similarly,  concealing the whole mouth results in a much more conventional
reconstruction of the mouth.  As a result, the residual error emphasizes where the image is anomalous.  For the second row, note how eye size and gaze are affected; and the significant
change in reconstructed mouth shape when the mouth is concealed.  This effect is minor for typical faces.  As a result, residuals against autoencoder inpainting are strong cues to anomaly.
}  
\label{fig:inpaint_AE_idea}
\end{figure*}

We view anomaly detection as feature construction followed by a simple unsupervised method. 
Natural choices of feature constructions are autoencoder codes, pretrained discriminative models (eg \cite{cao2017vggface2}), or autoencoder residual features.
An anomalous face image will look mostly like a typical face image, but will display some crucial differences.  The
problem is we don't know where those differences are or what they look like.   A natural strategy is based on a
generative models of typical face images.  Write $Q$ for a test image, and ${\cal M}(Q;\theta)$ for a learned model that
produces the typical face image that is `closest' to the query image.  
We could then use the difference ${\cal   M}(Q;\theta)-Q$ to compute a score of anomaly.   
In practice, ``peeking'' by the learned model (details below) means that this approach fails.  The learning
procedure results in a model that is biased to produce a ${\cal M}(Q;\theta)$ that is closer to $Q$ than it should
be.  

A simple variant of this approach is extremely effective.  Rather than requiring ${\cal M}(.; \theta)$ to make the closest
typical image, we conceal part of $Q$ from ${\cal M}(.;\theta)$ and require it to extrapolate.  We then compare the
extrapolated region to $Q$ to produce the anomaly signal.

\subsection{Autoencoder Residuals as Anomaly Signals}

We will build ${\cal M}$ using an autoencoder.  Autoencoders construct
low dimensional latent variable models from high dimensional signals. 
An encoder $\mathcal{E}$ estimates the latent variable
(code; $z$) for a given input $Q$; a decoder $\mathcal{D}$ recovers
the signal from that code.  The two are trained together, using criteria like the accuracy of the signal recovery (ie
$\vert \mathcal{D}(\mathcal{E}(Q))-Q\vert^2$;~\cite{bengio2009learning}).  Variational versions which use Bayes priors 
on the code have been explored as well~\cite{Kingma:2013tz}.
As we show in table~\ref{tbl:glasses}, the code produced by the encoder is a poor guide to anomaly, likely because it is still fairly high
in dimension, and an appropriate distribution model is obscure \cite{schlegl:2017au}.  The autoencoder image reconstruction residual,
$\mathcal{D}(\mathcal{E}(Q))-Q$, is an alternative. 

Straightforward experiments establish that the residual is a poor anomaly signal (Figure~\ref{fig:inpaint_AE_idea}). The reason is interesting.
An autoencoder is trained to reproduce signals from its training set, but this regime does not necessarily discourage
reproducing other images as well.  An autoencoder that is trained to reproduce face images accurately has not been
trained {\em not} to reproduce (say) cat images accurately, too.  This means the autoencoder could reduce the training
loss by adopting a compression strategy that works for many kinds of images.  Therefore, a compression procedure that is good
at compressing face images is not necessarily bad at compressing other images. 
This problem is not confined to neural networks. For example, choice of principal components that represents face images well~\cite{sirovich1987low} 
may represent (say) cat images. Denoising in current implementations~\cite{Vincent:2010vu} does not cure this problem.  For example, a good denoising
strategy is to construct a large dictionary of patches, then report the closest patch to the input.  While a dictionary built on faces may reproduce
some classes of image poorly, there is no guarantee in the training loss. Requiring a `small' code~\cite{hinton1994autoencoders}
or adding code regularization~\cite{Kingma:2013tz} does not cure this problem either, because it is
not known how to account for the information content of the code.  As a result, the model ${\cal M}$ built by the
autoencoder is not guaranteed to report the typical face image that is `closest' to the query image; instead, it may
pass through some of the query image as well (`peeking' at the query image), so resulting in a small residual and a
poor anomaly signal.   Experimental experience suggests that neural networks quite reliably adopt unexpected strategies
for minimizing loss (`cheating' during training), meaning that we expect peeking to occur, and figure~\ref{fig:inpaint_AE_idea} confirms that it does.
Peeking can be overcome by forcing the autoencoder to fill in large holes in the query image.
\section{Beating Peeking with Inpainting}
Write $\Pi_b$ for an operator that takes an image and overwrites a box $b$ with zeros; write $\Pi_{\overline b}$ for an
operator that overwrites all but the box $b$ with zeros.   These boxes will be quite large
in practice.   We will train an autoencoder $({\cal E}, {\cal D})$ as below.  We build an anomaly feature vector by
constructing $\vert \Pi_{\overline b}\left[{\cal D}({\cal E} (\Pi_b(Q)))-Q\right]\vert$ for a variety of boxes (as below).  We
will then apply simple decision procedures to this feature.

This feature works because the autoencoder cannot peek into $Q$ within the box.  Instead, it must extrapolate into $b$,
and that extrapolation is difficult to produce for multiple classes.  In turn, the extrapolate is a much better estimate
of what a typical face image would look like within $b$, conditioned on the rest of $Q$.  For example, if $b$ spans a
mouth, then the auto encoder constructs a typical mouth conditioned on the face and compares it with the observed mouth, and
if the mouth is anomalous the residual will be large (Figure~\ref{fig:inpaint_AE_idea}).

This is similar to the inpainting problem explored in~\cite{Context:CktiXfRg}, though we do not use an adversarial loss.
The autoencoder is trained to inpaint randomly selected boxes.  We use $\vert ({\cal D}({\cal E} (\Pi_b(Q)))-Q)\vert_1$ as a
training loss, thus requiring the autoencoder to inpaint.  The only difference between this and a denoising regime is the size of
the boxes, which is large compared to gaussian noise.

Our encoder and decoder use standard convolutional architectures with a fully connected layer for code construction.
Average pooling is used for downsampling, and bilinear interpolation is used for upsampling.  Following
\cite{berthelot2017began}, we use a higher capacity network for the encoder than the decoder which seems to help with
reconstructing higher frequency information.  We use the elu non-linearity and batch normalization after each conv layer,
and a tanh non-linearity on the output from the decoder.  We use the $L_1$ norm for our training loss.

\section{Eyeglass Experiment}
\label{sec:eyeglass}
\begin{table}[]
\centering
\begin{tabular}{llll}
\cline{2-4}
\multicolumn{1}{l|}{}         & \multicolumn{3}{c|}{$L_1$ Regularized Logistic Regressor}                                                 \\ \hline
\multicolumn{1}{|l|}{Feature} & \multicolumn{1}{l|}{Resnet50}      & \multicolumn{1}{l|}{AE Code} & \multicolumn{1}{l|}{Res Patch}     \\ \hline
\multicolumn{1}{|l|}{Accu}     & \multicolumn{1}{l|}{\textbf{99.4}} & \multicolumn{1}{l|}{85.6}    & \multicolumn{1}{l|}{93.5}          \\ \hline
                              &                                    &                              &                                    \\ \cline{2-4} 
\multicolumn{1}{l|}{}         & \multicolumn{3}{c|}{1-Class SVM}                                                                       \\ \hline
\multicolumn{1}{|l|}{Feature} & \multicolumn{1}{l|}{Resnet50}      & \multicolumn{1}{l|}{AE Code} & \multicolumn{1}{l|}{Res Patches}   \\ \hline
\multicolumn{1}{|l|}{AUC}     & \multicolumn{1}{l|}{52.3}          & \multicolumn{1}{l|}{51.8}    & \multicolumn{1}{l|}{53.8}          \\ \hline
                              &                                    &                              &                                    \\ \cline{2-4} 
\multicolumn{1}{l|}{}         & \multicolumn{3}{c|}{Mahalanobis Distance}                                                              \\ \hline
\multicolumn{1}{|l|}{Feature} & \multicolumn{1}{l|}{Resnet50}      & \multicolumn{1}{l|}{AE Code} & \multicolumn{1}{l|}{Res Patches}   \\ \hline
\multicolumn{1}{|l|}{AUC}     & \multicolumn{1}{l|}{52.8}          & \multicolumn{1}{l|}{50.8}    & \multicolumn{1}{l|}{\textbf{92.5}} \\ \hline
                              &                                    &                              &                                    \\ \cline{2-4} 
\multicolumn{1}{l|}{}         & \multicolumn{3}{c|}{$L_\infty$ norm}                                                                   \\ \hline
\multicolumn{1}{|l|}{Feature} & \multicolumn{1}{l|}{Resnet50}      & \multicolumn{1}{l|}{AE Code} & \multicolumn{1}{l|}{Res Patches}   \\ \hline
\multicolumn{1}{|l|}{AUC}     & \multicolumn{1}{l|}{NA}            & \multicolumn{1}{l|}{NA}      & \multicolumn{1}{l|}{\textbf{81.5}} \\ \hline
\end{tabular}
\caption{Supervised and unsupervised glasses detection with our features.  We compare our inpainting residual features, \textbf{Res patch} to codes from the autoencoder and state of the art Resnet features trained on face detection.  We note that Resnet features work extremely well for supervised tasks, in fact in our simple dataset, Resnet features combined with an $L_1$ regularized logistic regression performs nearly perfectly.  However Resnet features do not perform well for either of the unsupervised classifiers.  On the other hand, the inpainting residual features perform better than the autoencoder code regardless of the classifier type, and perform far better than Resnet for unsupervised classifiers.}
\label{tbl:glasses}
\vspace{-15pt}
\end{table}

Poor feature performance on a supervised task suggests that unsupervised methods will perform poorly too.  Therefore in evaluating feature constructions, 
it can be useful to compare to an oracle.  To do so, we construct a proxy anomaly experiment, where anomalous faces
are those wearing eyeglasses, so allowing discriminative training of the oracle.  Our oracle takes the form of a supervised $L_1$-regularized linear regressor 
(we use glmnet~\cite{glmnet}) trained on data.  While poor performance on the oracle suggests that unsupervised methods will perform poorly too, 
good performance on the oracle is not necessarily indicative of good unsupervised performance. We therefore also explore natural choices 
for unsupervised methods including one-class SVM~\cite{scholkopf2000support}, one-class density estimates such as Mahalanobis Distance~\cite{mahalanobis1936generalized}, 
and heuristic methods such as the $L_\infty$ norm which are meaningful for our residual based feature.

We use the Celeb-A dataset \cite{liu2015faceattributes}, which is a collection of thousands of labeled faces. As in~\cite{berthelot2017began}, 
we filter and crop with the Viola-Jones face detector~\cite{viola2001rapid}, resulting in frontal faces in tightly cropped 128x128 boxes.
For this experiment, we use 7700 images of people wearing eyeglasses as anomalies and 7000 images without eyeglasses as our unsupervised training set. 
We train our inpainting autoencoder with random $\Pi_b$ for each sample on 90\% of the non-eyeglass data.  During test we use the same model to construct autoencoder codes 
as well as the inpainting features.  For inpainting features, we use 32x32 boxes in a regular grid.  We exclude the boxes that would lie directly on the image boundary. 
For Resnet features we use a pretrained resnet trained on face recognition from \cite{cao2017vggface2}, we remove the final softmax layer, and use the resulting network as a 
feature constructor.

Our results shown in table~\ref{tbl:glasses} suggest that inpainting autoencoder residuals contain sufficient information for attribute classification.  
One class SVM's are not a strong baseline for this problem.  Mahoanobis distance is a decent baseline for our features. 
Inpainting autoencoder residual images are informative, even with the simplest heuristic classifier $L_\infty$. 
We also show that autoencoder codes are less adept than inpainting residuals.

It is not surprising that the $L_\infty$ classifier works so well.  Each inpainting feature is a local anomaly detector for the content
under the box.  Since the inpainting autoencoder was trained on images without glasses, when the box covers the eyeglasses, the
inpanted content will not have eyeglasses.  Consequently, the residual will be large.  Taking the $L_\infty$ norm reports the score from the most violated box.  
Note that this experiment is not a true anomaly experiment, but it is similar to previous work~\cite{zhai2016deep, kliger2018novelty,deecke2018anomaly,gu2018semisupervised} 
which uses attributes or class labels as proxies for anomalies.

\section{Anomaly Experiment}
\label{sec:trueanom}
\begin{figure}[t]
\centering
\begin{subfigure}[b]{.3\linewidth}
\includegraphics[width=\linewidth]{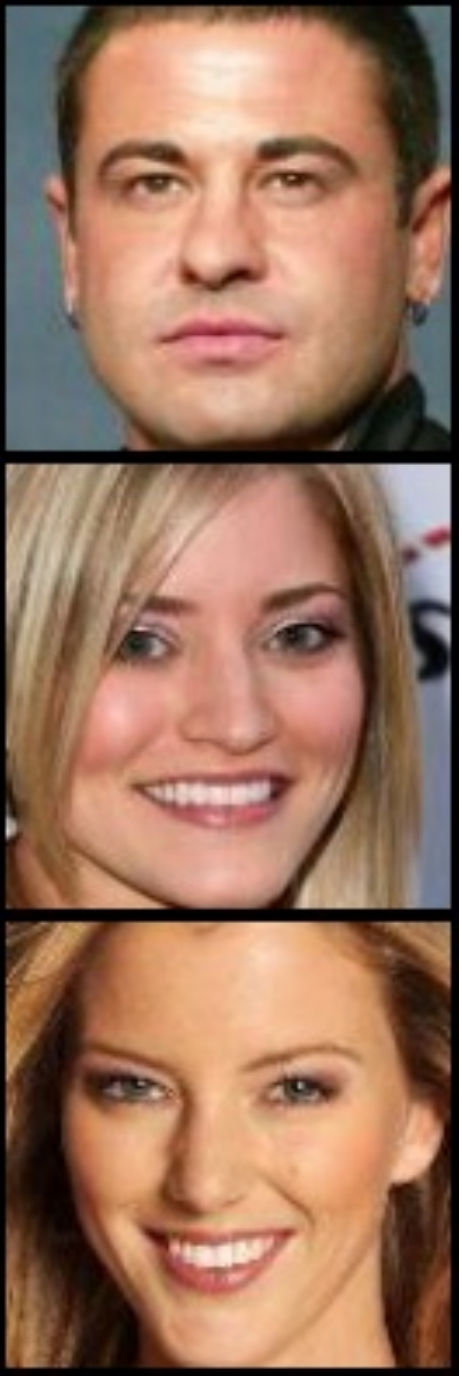}
\caption{Celeb A}
\label{subfig:celebA}
\end{subfigure}
\begin{subfigure}[b]{.3\linewidth}
\includegraphics[width=\linewidth]{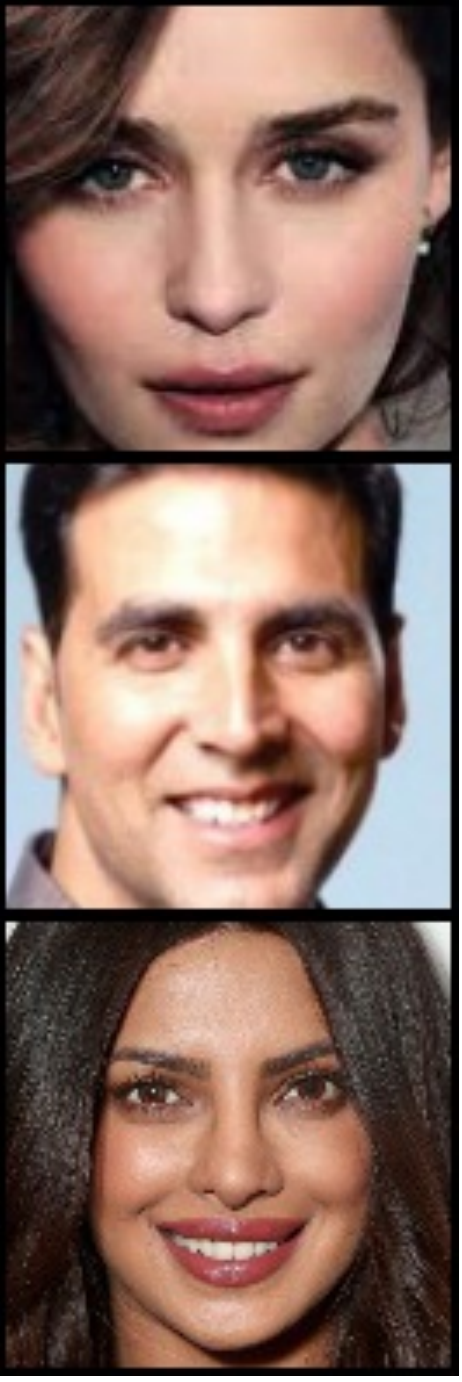}
\vspace{-10pt}
\caption{Typical Set}
\label{subfig:ordinary}
\end{subfigure}
\begin{subfigure}[b]{.3\linewidth}
\includegraphics[width=\linewidth]{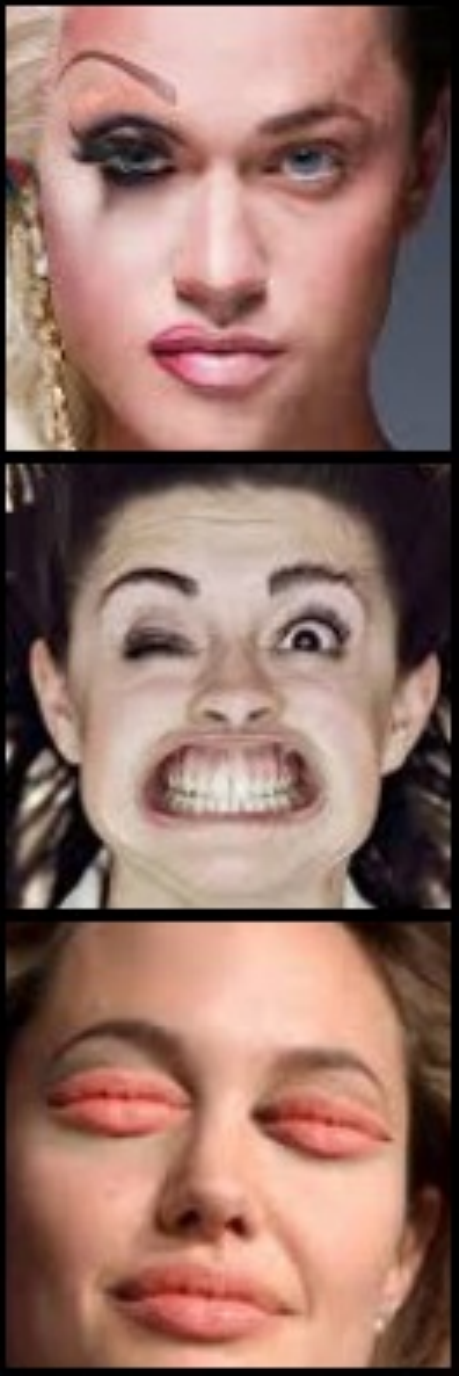}
\vspace{-10pt}
\caption{Anomaly Set}
\label{subfig:anomaly}
\end{subfigure}
\caption{Images from celeb A \ref{subfig:celebA}, from the \textbf{typical set} \ref{subfig:ordinary} and from the \textbf{anomaly set} \ref{subfig:anomaly}.
  It is very clear when an image is an anomaly and when it is not. Our typical set is similar to celeb-A with recent new images collected after celeb-A was created, 
  ensuring no overlaps with celeb-A images. However, there are some slight features of Celeb-A images that are
  noticeably different from our typical set.  For example, the resolution of the images seems to be slightly different.}
\label{fig:anomaly_examples}
\vspace{-5pt}
\end{figure}

We wish to determine whether we can detect true anomalies in a realistic setting.  We use the Celeb-A dataset as our training data, filtering and cropping for frontal faces using Viola-Jones as in section~\ref{sec:eyeglass}.  However rather than considering any specific attribute, we consider the entire Celeb-A dataset as typical data.  We set aside 30,000 images for use in test, leaving us with 125,253 images for training. For our anomaly images, we collect a 100 image anomaly dataset.  This set, which we call the \textbf{anomaly set} is comprised of strange or ``weird'' faces (Figure~\ref{subfig:anomaly}). It includes extreme makeup, masks, photoshops, and people making extreme faces.
We pass the images through the same Viola-Jones detector and cropper. This rejects many of the anomaly images,
and in fact we only have about a 3\% yield on anomaly images, meaning that we had to find roughly 3000 anomaly images in order to get 100. 
However, this construction is sensible: if Viola-Jones does not believe the images have a face, then they are too obviously anomalous. 
For example, a photograph of a cat would have a high anomaly score under our method, but a cat is also not likely to be identified as a face by a competent detector, 
so determining it is an anomaly is not particularly important or difficult.

We also wish to determine if anomaly detection is caused by special features of the Celeb-A dataset.  An anomaly detector which identifies any image not from Celeb-A 
would not be particularly useful. We therefore collect a \textbf{typical set} of 100 images we do not believe to be anomalous images. It is comprised of pictures of celebrities 
that were taken after Celeb-A was created so there are no overlaps in pictures (Figure~\ref{subfig:ordinary}).  We also tried to find new celebrities, so that the people would be 
less likely to have appeared in the original Celeb-A dataset.  This dataset is used to validate that a method is not memorizing images in Celeb-A or finding a particular feature of 
Celeb-A and rejecting any new images.

We show samples from Celeb-A, the typical set and the anomaly set in figure~\ref{fig:anomaly_examples}.  By example it is reasonable to ask an anomaly detection method to 
identify images from the anomaly set without identifying images from the typical set. 

Our experiments are modeled on the set experiment presented in \cite{Zaheer:2017wn}.  They form a set of 16 images from Celeb-A  where 15 images share at two attributes, 
and one image differs. The goal is to identify the image with different attributes.  We adjust this slightly.  Large sets are more indicative of performance for real world 
anomaly detection, where the goal is to identify one image in thousands rather than one image in ten.  However, using large sets is significantly more difficult so we report 
recall at 1, 5, and 10 rather than just reporting recall at 1.  Note that at no point does any method have access to labels, which are revealed only to evaluate the experiment.  
We believe this is a better model for detecting rare anomalies.

\begin{figure*}
\centering
\vspace{-5pt}
  \begin{subfigure}[b]{\linewidth}
    \centering
    \includegraphics[width=\linewidth]{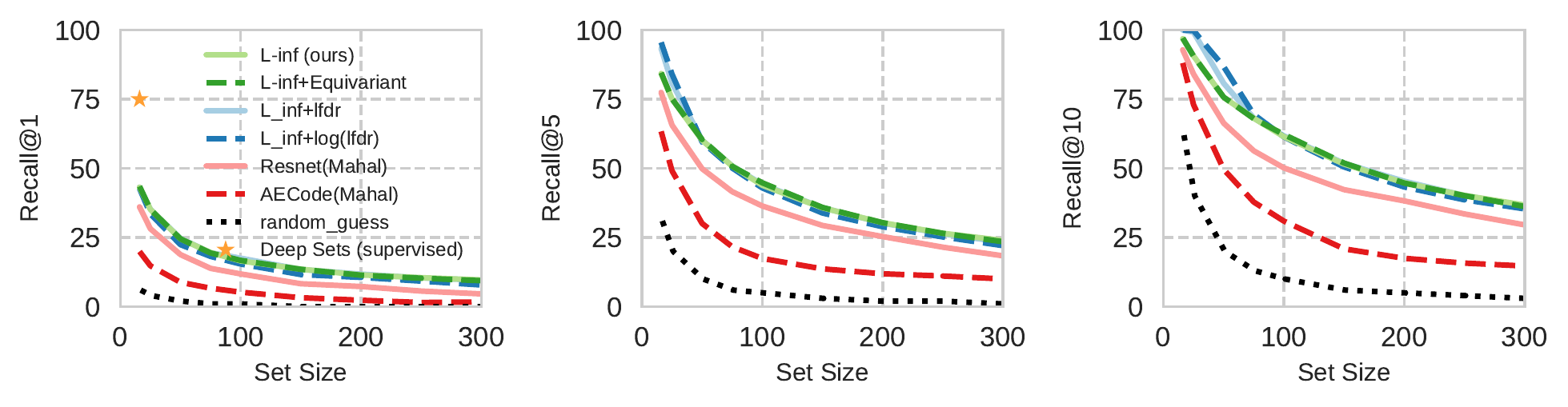}
    \caption{Recall of images from the \textbf{anomaly set}}
  \end{subfigure}
    \centering
    \begin{subfigure}[b]{\linewidth}
     \includegraphics[width=\linewidth]{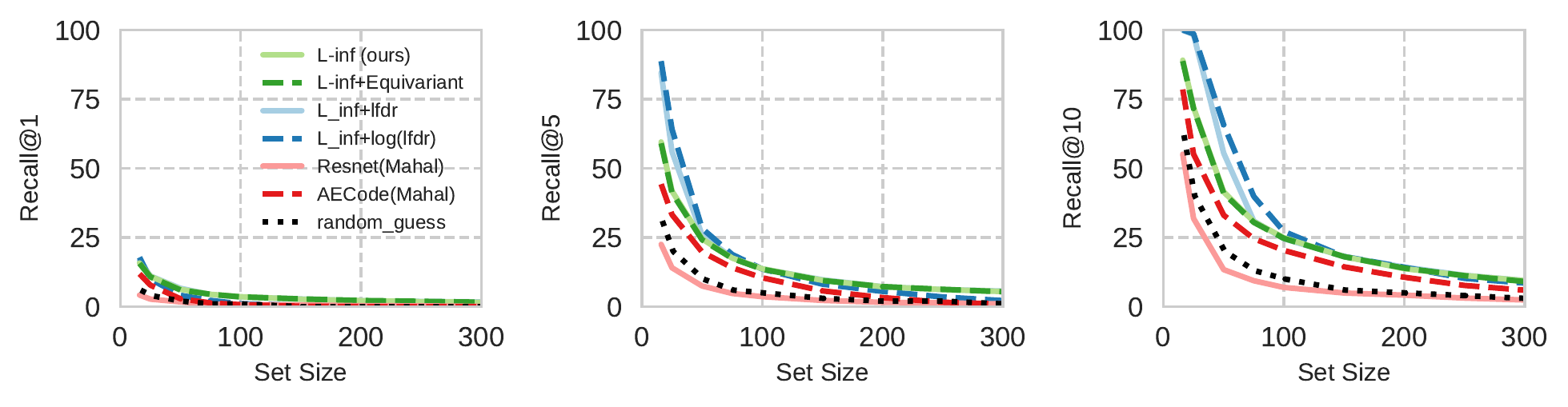}
    \caption{Recall of images from the \textbf{typical set}}
  \end{subfigure}
\caption{{\bf Top row:}  Recall at 1, 5, and 10 for various scores using our anomaly feature plotted against the size of the set from which the anomaly must be picked. We show the performance on our new data set for $L_\infty$ on the autoencoders' inpainted features, adding equivariant transformation to our features, local false discovery rate (lfdr and $log$(lfdr)), and Mahalanobis distance on features drawn from Resnet50 and autoencoder's code as our baseline.
Recalls are averaged over 10000 trials, and so have very low variance.  Note there is very little visible difference between the performance of the scores, all
of which beat chance very strongly. The equivariant transformations does not have any major effect to the performance and our feature construction method in itself encodes a strong representation of anomaly. The star in the Recall@1 figure is reproduced from the work of Deep Sets \cite{Zaheer:2017wn}. Though, there is no direct comparison of our method to Deep Sets, the difference in performance is an indicator of the gap between supervised and unsupervised methods. Note also the test is demanding compared to the literature; a single anomalous face must be picked from up to 300 others.
However, these results might depend on some signal property of the celeb-A dataset.  
  {\bf Bottom row} shows results from our control experiment, where the image used as an anomalous image is a typical face image that doesn't appear in celeb-A (details in section~\ref{sec:trueanom}).
Performance is not at chance (suggesting that celeb-A images have some hitherto not noted special properties) but is close.  In particular, the performance of the
anomaly detector on anomalous images very strongly exceeds its performance on control images, and so cannot be explained by the special properties of celeb-A (whatever they are).}  
\vspace{-25pt}
\label{fig:quant_anomalies}
\end{figure*}

{\bf Evaluating anomaly detection:}  We select one image from the anomaly set, and
between 15 and 299 images from the 30,000 celeb-A held out images (without consideration of attributes, in contrast to~\cite{Zaheer:2017wn}). 
We then score each image using our feature and a variety of scoring methods (section~\ref{sec:unsup_feat_learn}) 
to evaluate recall for the anomaly image, averaged over 10,000 sets. 
As figure~\ref{fig:quant_anomalies} shows, recall is strong even from large sets, and the choice of score appears not to matter.

{\bf Control:}  Strong results could be caused by some special feature of celeb-A images.  To control
for this possibility, we repeat the anomaly detection experiment, but replacing the image from the anomaly set with an image
from the typical set (100 typical images not from celeb-A).  If celeb-A were wholly representative, then
this experiment should produce recalls at chance.  As figure~\ref{fig:quant_anomalies} shows, the results are not at chance
(there is something interesting lurking in celeb-A), but recall is very much weaker than for anomalies.
The performance of the anomaly detector cannot be explained by quirks of celeb-A
\begin{figure*}
\centering
\includegraphics[width=\linewidth]{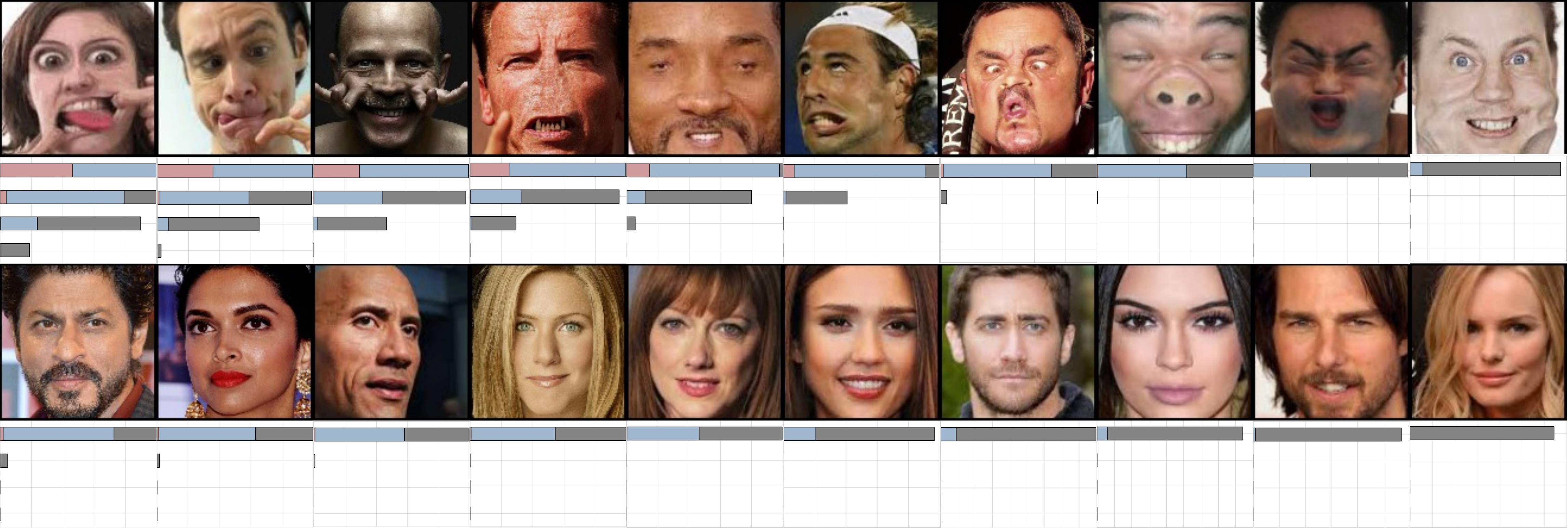}
\vspace{-15pt}
\caption{\textbf{Top row:} images from the anomaly set sorted by their $L_\infty$ anomaly scores. The median image from each decile.
\textbf{Bottom row:} images from the ordinary set sorted by their $L_\infty$ anomaly scores.  The median image from each decile is shown.
Bar charts below show how frequently the image
was identified @1 (red), @5 (blue), and @10 (gray) for (top to bottom) 16, 64, 128, and 256 image sets. For anomaly images (top row),
being identified frequently is better, for ordinary images (bottom row) being identified less is better.}
\label{fig:qualitative_anomalies}
\end{figure*}

\subsection{Unsupervised Feature Learning}
\label{sec:unsup_feat_learn}
We use our regular grid of residual features for 32x32 patches with a 32 pixel edge exclusion and explore a variety of methods 
for turning the residual features into an anomaly score. The $L_\infty$ norm over the feature vector makes up our main method due to its simplicity and good performance. 
It is not obvious that this is a good choice, and for a general feature, this norm would be largely meaningless.  
However, as demonstrated in the attribute classification task our features are designed to be well suited to this norm. 
For our feature, the $L_\infty$ norm finds the most violated residual from the set of patches, 
which is obviously useful for anomalies that tend to occur locally.

The \textbf{Mahalanobis Distance} (mahal) estimates a mean and covariance from a set and then measures distance with respect to the mean and covariance. 
It is typical to estimate the mean and covariance on training data.The \textbf{Equivariant Transform} (equivariant) introduced in \cite{Zaheer:2017wn} 
can be applied in an unsupervised manner on a set of images. A sensible version looks like the Mahalanobis Distance. Recall the equivariant transform in matrix form:
\begin{equation}
\mathbf{X} = \left[\lambda \mathbf{I} + \gamma (\mathbf{11^T})\right] \mathbf{X}
\end{equation}
Which for an element $x_i$ is equivalent to
\begin{equation}
\hat{x_i} = \lambda x_i + \gamma \sum_i x_i
\vspace{-5pt}
\end{equation}

Let $\Sigma^{-1}$ be the inverse covariance of $\mathbf{X}$, 
then $\gamma = -\Sigma^{-1/2}/N$ and $\lambda = \Sigma^{-1/2}$ compute a transformation which under the $L_2$ 
is the Mahalanobis Distance. This transformation is sensible and can be applied to the data prior to applying the $L_\infty$ norm to reweight
the feature dimensions and take into account that some dimensions might be highly varying while others are not. 
For our autoencoder residual feature, we assume that our features are IID, so we can estimate a diagonal covariance, and we compute a robust mean and covariance 
by eliminating the largest and smallest values on each feature.  Note that this is done without knowing which item is anomalous and thus does not violate train-test splits.

The {\bf Local False Discovery Rate} (lfdr) is a construction that identifies the probability that an item comes from a null distribution, 
without knowing what the null is~\cite{Efron2007}. The method originates in multiple hypothesis testing, assuming that most observations come from the null. 
Assume the null distribution is $f_o(z)$, the non-null is $f_1(z)$, and the prior an item comes from the null is $\pi_o$.  Then the lfdr is
\begin{equation}
p(\mbox{null}|z)=\frac{\pi_o f_o(z)}{\pi_o f_o(z)+(1-\pi_o) f_1(z)}
\end{equation}
Small values suggest an item is worth investigating (i.e., anomalous).  Estimation is complicated by the fact that neither $f_o(z)$ nor $f_1(z)$ are known; but the assumption that
$\pi_o$ is large, and $f_o(z)$ is `close' to a standard normal distribution allows fairly accurate estimation.  We used the R program {\tt locfdr}.  
We estimated local false discovery rates using all 30200 test data items (doing so does not involve knowing which item is anomalous, so does not violate test-train protocols). 
We estimate using a standardized version of the L-infinity score, and a standardized version of the log of the L-infinity score.

\subsection{Results}
As seen in figure~\ref{fig:quant_anomalies}, our feature performs well regardless of feature transformation applied. 
We report performance from \cite{Zaheer:2017wn} on the graph, even though their experiment is on different data.  
While they outperform our method for 16 image sets, using our auto-encoder residual features, we identify \textbf{anomalies} 
at rates significantly greater than chance even as the size of the set increases. Resnet-50 features \cite{cao2017vggface2} with a 
Mahalanobis Distance represents a strong baseline, however, we outperform it. There does seem to be some bias in the Celeb-A dataset 
being used to identify anomalies but our features and the Resnet-50 features do not identify \textbf{typical} images at anywhere near the same rate as anomalies.  
The gap between performance on typical images and anomalous images is apparent and clearly significant (eg. 40 vs 20 percent for recall at 1 in a 16 image set).

For qualitative comparison, we show the median image from each decile ranked by their $L_\infty$ anomalous score in figure~\ref{fig:qualitative_anomalies}.  
We also show a plot of how frequently they are ranked in the top images in a set of increasing size. Anomaly images are frequently identified as the most anomalous image 
in a 16 image set and as a top 10 anomalous image in 128 image sets.  Typical images are almost never identified as the most anomalous image in any set, 
and almost never identified as the top 10 in any set larger than 16.  The median least anomalous image is roughly as anomalous as the median typical image.  
These findings are consistent with our quantitative results, which show that images from the anomalous set are identified frequently and images from the typical set are 
identified more often than chance, but frequently less often than true anomalies.

\section{Conclusion}
We introduce the inpainting autoencoder residual as a feature for combating the overgeneralization of compression losses. 
This allows us to train our method solely on non-anomalous data, mimicking how a real anomaly detector must be trained.  
We demonstrate that our inpainting residual features are useful and work well in supervised and unsupervised settings. 
Though we did not see improvement in performance, it is easy to use inpainting autoencoder features with various feature transformation techniques.  
We also describe a standard anomaly detection experiment for evaluating future anomaly work on image sets, enabled through the collect two small datasets to augment Celeb-A.

\bibliography{anomaly}

\begin{thebibliography}{31}
\providecommand{\natexlab}[1]{#1}
\providecommand{\url}[1]{\texttt{#1}}
\expandafter\ifx\csname urlstyle\endcsname\relax
  \providecommand{\doi}[1]{doi: #1}\else
  \providecommand{\doi}{doi: \begingroup \urlstyle{rm}\Url}\fi

\bibitem[Alpert \& Kisilev(2014)Alpert and Kisilev]{Alpert:2014bz}
Alpert, Sharon and Kisilev, Pavel.
\newblock {Unsupervised detection of abnormalities in medical images using
  salient features}.
\newblock In Ourselin, Sebastien and Styner, Martin~A (eds.), \emph{SPIE
  Medical Imaging}, pp.\  903416--7. SPIE, March 2014.

\bibitem[Arashloo et~al.(2017)Arashloo, Kittler, and
  Christmas]{Arashloo:2017ge}
Arashloo, Shervin~Rahimzadeh, Kittler, Josef, and Christmas, William.
\newblock {An Anomaly Detection Approach to Face Spoofing Detection: A New
  Formulation and Evaluation Protocol}.
\newblock \emph{IEEE Access}, 5:\penalty0 13868--13882, 2017.

\bibitem[Bengio et~al.(2009)]{bengio2009learning}
Bengio, Yoshua et~al.
\newblock Learning deep architectures for ai.
\newblock \emph{Foundations and trends{\textregistered} in Machine Learning},
  2\penalty0 (1):\penalty0 1--127, 2009.

\bibitem[Berthelot et~al.(2017)Berthelot, Schumm, and Metz]{berthelot2017began}
Berthelot, David, Schumm, Tom, and Metz, Luke.
\newblock Began: Boundary equilibrium generative adversarial networks.
\newblock \emph{arXiv preprint arXiv:1703.10717}, 2017.

\bibitem[Cao et~al.(2017)Cao, Shen, Xie, Parkhi, and
  Zisserman]{cao2017vggface2}
Cao, Qiong, Shen, Li, Xie, Weidi, Parkhi, Omkar~M, and Zisserman, Andrew.
\newblock Vggface2: A dataset for recognising faces across pose and age.
\newblock \emph{arXiv preprint arXiv:1710.08092}, 2017.

\bibitem[Chandola et~al.(2009)Chandola, Banerjee, and Kumar]{Chandola:2009fo}
Chandola, Varun, Banerjee, Arindam, and Kumar, Vipin.
\newblock {Anomaly Detection: A Survey}.
\newblock \emph{Acm Computing Surveys}, 41\penalty0 (3), 2009.

\bibitem[Deecke et~al.(2018)Deecke, Vandermeulen, Ruff, Mandt, and
  Kloft]{deecke2018anomaly}
Deecke, Lucas, Vandermeulen, Robert, Ruff, Lukas, Mandt, Stephan, and Kloft,
  Marius.
\newblock Anomaly detection with generative adversarial networks, 2018.
\newblock URL \url{https://openreview.net/forum?id=S1EfylZ0Z}.

\bibitem[Deshpande et~al.(2017)Deshpande, Lu, Yeh, Chong, and
  Forsyth]{Deshpande:2017ik}
Deshpande, Aditya, Lu, Jiajun, Yeh, Mao-Chuang, Chong, Min~Jin, and Forsyth,
  David.
\newblock {Learning Diverse Image Colorization}.
\newblock In \emph{2017 IEEE Conference on Computer Vision and Pattern
  Recognition (CVPR)}, pp.\  2877--2885. IEEE, 2017.

\bibitem[Efron(2007)]{Efron2007}
Efron, Bradley.
\newblock Size, power and false discovery rates.
\newblock \emph{Ann. Statist.}, 35\penalty0 (4):\penalty0 1351--1377, 08 2007.
\newblock \doi{10.1214/009053606000001460}.
\newblock URL \url{https://doi.org/10.1214/009053606000001460}.

\bibitem[Friedman et~al.(2010)Friedman, Hastie, and Tibshirani]{glmnet}
Friedman, Jerome, Hastie, Trevor, and Tibshirani, Robert.
\newblock Regularization paths for generalized linear models via coordinate
  descent.
\newblock \emph{Journal of Statistical Software}, 33\penalty0 (1):\penalty0
  1--22, 2010.
\newblock URL \url{http://www.jstatsoft.org/v33/i01/}.

\bibitem[Goodfellow et~al.(2014)Goodfellow, Pouget-Abadie, Mehdi, Xu,
  Warde-Farley, Ozair, Courville, and Bengio]{Goodfellow:2014wp}
Goodfellow, Ian~J, Pouget-Abadie, Jean, Mehdi, Mirza, Xu, Bing, Warde-Farley,
  David, Ozair, Sherjil, Courville, Aaron, and Bengio, Yoshua.
\newblock {Generative Adversarial Networks}.
\newblock In \emph{NIPS}, June 2014.

\bibitem[Hasler et~al.(2003)Hasler, Sbaiz, Susstrunk, and
  Vetterli]{Hasler:2003dz}
Hasler, D, Sbaiz, L, Susstrunk, S, and Vetterli, M.
\newblock {Outlier modeling in image matching}.
\newblock \emph{IEEE TPAMI}, 25\penalty0 (3):\penalty0 301--315, March 2003.

\bibitem[Hinton \& Salakhutdinov(2006)Hinton and
  Salakhutdinov]{hinton2006reducing}
Hinton, Geoffrey~E and Salakhutdinov, Ruslan~R.
\newblock Reducing the dimensionality of data with neural networks.
\newblock \emph{science}, 313\penalty0 (5786):\penalty0 504--507, 2006.

\bibitem[Hinton \& Zemel(1994)Hinton and Zemel]{hinton1994autoencoders}
Hinton, Geoffrey~E and Zemel, Richard~S.
\newblock Autoencoders, minimum description length and helmholtz free energy.
\newblock In \emph{Advances in neural information processing systems}, pp.\
  3--10, 1994.

\bibitem[Jindong~Gu \& Tresp(2018)Jindong~Gu and Tresp]{gu2018semisupervised}
Jindong~Gu, Matthias~Schubert and Tresp, Volker.
\newblock Semi-supervised outlier detection using generative and adversary
  framework, 2018.
\newblock URL \url{https://openreview.net/forum?id=BkS3fnl0W}.

\bibitem[Kingma \& Welling(2014)Kingma and Welling]{Kingma:2013tz}
Kingma, Diederik~P and Welling, Max.
\newblock {Auto-Encoding Variational Bayes}.
\newblock In \emph{ICLR}, 2014.

\bibitem[Kliger \& Fleishman(2018)Kliger and Fleishman]{kliger2018novelty}
Kliger, Mark and Fleishman, Shachar.
\newblock Novelty detection with {GAN}, 2018.
\newblock URL \url{https://openreview.net/forum?id=Hy7EPh10W}.

\bibitem[Lample et~al.(2017)Lample, Zeghidour, Usunier, Bordes, Denoyer, and
  Ronzato]{Lample:2017tw}
Lample, Guillaume, Zeghidour, Neil, Usunier, Nicolas, Bordes, Antoine, Denoyer,
  Ludovic, and Ronzato, MarcAurelio.
\newblock {Fader Networks: Manipulating Images by Sliding Attributes}.
\newblock \emph{arXiv.org}, pp.\  1--10, June 2017.

\bibitem[Liu et~al.(2015)Liu, Luo, Wang, and Tang]{liu2015faceattributes}
Liu, Ziwei, Luo, Ping, Wang, Xiaogang, and Tang, Xiaoou.
\newblock Deep learning face attributes in the wild.
\newblock In \emph{Proceedings of International Conference on Computer Vision
  (ICCV)}, December 2015.

\bibitem[Mahalanobis(1936)]{mahalanobis1936generalized}
Mahalanobis, Prasanta~Chandra.
\newblock On the generalized distance in statistics.
\newblock National Institute of Science of India, 1936.

\bibitem[Mak et~al.(2005)Mak, Peng, and Lau]{Mak:2005gy}
Mak, K~L, Peng, P, and Lau, H Y~K.
\newblock {A real-time computer vision system for detecting defects in textile
  fabrics}.
\newblock In \emph{2005 IEEE International Conference on Industrial
  Technology}, pp.\  469--474. IEEE, 2005.

\bibitem[Pathak et~al.(2016)Pathak, Kr{\"a}henb{\"u}hl, Donahue, Darrell, and
  Efros]{Context:CktiXfRg}
Pathak, Deepak, Kr{\"a}henb{\"u}hl, Philipp, Donahue, Jeff, Darrell, Trevor,
  and Efros, Alexei~A.
\newblock {Context Encoders: Feature Learning by Inpainting}.
\newblock In \emph{Computer Vision and Pattern Recognition}, April 2016.

\bibitem[Schlegl et~al.(2017)Schlegl, Seeb{\"o}ck, Waldstein, Schmidt-Erfurth,
  and Langs]{schlegl:2017au}
Schlegl, Thomas, Seeb{\"o}ck, Philipp, Waldstein, Sebastian~M, Schmidt-Erfurth,
  Ursula, and Langs, Georg.
\newblock Unsupervised anomaly detection with generative adversarial networks
  to guide marker discovery.
\newblock In \emph{International Conference on Information Processing in
  Medical Imaging}, pp.\  146--157. Springer, 2017.

\bibitem[Sch{\"o}lkopf et~al.(2000)Sch{\"o}lkopf, Williamson, Smola,
  Shawe-Taylor, and Platt]{scholkopf2000support}
Sch{\"o}lkopf, Bernhard, Williamson, Robert~C, Smola, Alex~J, Shawe-Taylor,
  John, and Platt, John~C.
\newblock Support vector method for novelty detection.
\newblock In \emph{Advances in neural information processing systems}, pp.\
  582--588, 2000.

\bibitem[Serdaroglu et~al.(2006)Serdaroglu, Ertuzun, and
  Ercil]{Serdaroglu:2006he}
Serdaroglu, A, Ertuzun, A, and Ercil, A.
\newblock {Defect detection in textile fabric images using wavelet transforms
  and independent component analysis}.
\newblock \emph{Pattern Recognition and Image Analysis}, 16\penalty0
  (1):\penalty0 61--64, 2006.

\bibitem[Sirovich \& Kirby(1987)Sirovich and Kirby]{sirovich1987low}
Sirovich, Lawrence and Kirby, Michael.
\newblock Low-dimensional procedure for the characterization of human faces.
\newblock \emph{Josa a}, 4\penalty0 (3):\penalty0 519--524, 1987.

\bibitem[Vincent et~al.(2010)Vincent, Larochelle, Lajoie, Bengio, and
  Manzagol]{Vincent:2010vu}
Vincent, Pascal, Larochelle, Hugo, Lajoie, Isabelle, Bengio, Yoshua, and
  Manzagol, Pierre-Antoine.
\newblock {Stacked Denoising Autoencoders: Learning Useful Representations in a
  Deep Network with a Local Denoising Criterion}.
\newblock \emph{The Journal of Machine Learning Research}, 11:\penalty0
  3371--3408, December 2010.

\bibitem[Viola \& Jones(2001)Viola and Jones]{viola2001rapid}
Viola, Paul and Jones, Michael.
\newblock Rapid object detection using a boosted cascade of simple features.
\newblock In \emph{Computer Vision and Pattern Recognition}, volume~1. IEEE,
  2001.

\bibitem[Yan et~al.(2016)Yan, Yang, Sohn, and Lee]{AttributeImageCon:Cqjisrcl}
Yan, Xinchen, Yang, Jimei, Sohn, Kihyuk, and Lee, Honglak.
\newblock {Attribute2Image: Conditional Image Generation from Visual
  Attributes}.
\newblock In \emph{European Conference on Computer Vision}, 2016.

\bibitem[Zaheer et~al.(2017)Zaheer, Kottur, Ravanbakhsh, Poczos, Salakhutdinov,
  and Smola]{Zaheer:2017wn}
Zaheer, Manzil, Kottur, Satwik, Ravanbakhsh, Siamak, Poczos, Barnabas,
  Salakhutdinov, Ruslan~R, and Smola, Alexander~J.
\newblock {Deep Sets}.
\newblock In \emph{NIPS}, pp.\  3394--3404, 2017.

\bibitem[Zhai et~al.(2016)Zhai, Cheng, Lu, and Zhang]{zhai2016deep}
Zhai, Shuangfei, Cheng, Yu, Lu, Weining, and Zhang, Zhongfei.
\newblock Deep structured energy based models for anomaly detection.
\newblock In \emph{International Conference on Machine Learning}, pp.\
  1100--1109, 2016.

\end{thebibliography}
\bibliographystyle{icml2018}

\end{document}